\documentclass[fleqn,10pt]{wlscirep}
\usepackage[utf8]{inputenc}
\usepackage[T1]{fontenc}
\usepackage{hyperref}
\hypersetup{
    colorlinks=true,
    linkcolor=blue,
    filecolor=magenta,      
    urlcolor=cyan,
    pdftitle={Overleaf Example},
    pdfpagemode=FullScreen,
    }
\newcommand{\termlmadda}{SIT-ADDA}

\title{Uncertainty-Guided Selective Adaptation Enables Cross-Platform Predictive Fluorescence Microscopy}

\author[1]{Kai-Wen K. Yang}
\author[2]{Andrew Bai}
\author[3, 4]{Alexandra Bermudez}
\author[2]{Yunqi Hong}
\author[4]{Zoe Latham}
\author[4]{Iris Sloan}
\author[2]{Michael Liu}
\author[2]{Vishrut Goyal}
\author[2*]{Cho-Jui Hsieh}
\author[3, 4, 5, 6*]{Neil Y.C. Lin}

\affil[1]{Independent Researcher, San Francisco, CA, USA}
\affil[2]{Computer Science Department, University of California, Los Angeles, CA, USA}
\affil[3]{Mechanical and Aerospace Engineering Department, University of California, Los Angeles, CA, USA}
\affil[4]{Bioengineering Department, University of California, Los Angeles, CA, USA}
\affil[5]{Jonsson Comprehensive Cancer Center, University of California, Los Angeles, CA, USA}
\affil[6]{Institute for Quantitative and Computational Biosciences, University of California, CA, USA}

\affil[*]{chohsieh@cs.ucla.edu, neillin@g.ucla.edu}

\begin{abstract}
Deep learning is transforming microscopy, yet models often fail when applied to images from new instruments or acquisition settings. Conventional adversarial domain adaptation (ADDA) retrains entire networks, often disrupting learned semantic representations. Here, we overturn this paradigm by showing that adapting only the earliest convolutional layers, while freezing deeper layers, yields reliable transfer. Building on this principle, we introduce Subnetwork Image Translation ADDA with automatic depth selection (SIT-ADDA-Auto), a self-configuring framework that integrates shallow-layer adversarial alignment with predictive uncertainty to automatically select adaptation depth without target labels. We demonstrate robustness via multi-metric evaluation, blinded expert assessment, and uncertainty–depth ablations. Across exposure and illumination shifts, cross-instrument transfer, and multiple stains, SIT-ADDA improves reconstruction and downstream segmentation over full-encoder adaptation and non-adversarial baselines, with reduced drift of semantic features. Our results provide a design rule for label-free adaptation in microscopy and a recipe for field settings; the code is publicly available.

\end{abstract}
\begin{document}

\flushbottom
\maketitle
%
%
\thispagestyle{empty}

\section*{Introduction}

Deep learning is reshaping optical bioimaging by redefining what microscopes can observe, resolve, and interpret. Neural networks increasingly act as algorithmic optics, revealing biological structures once buried in noise, blurred by diffraction, or invisible to conventional contrast~\cite{guo2025deep, lu2024diffusion, priessner2024content, imboden2021investigating}. This transformation spans a continuum: from recovering lost signal, to transcending optical limits, to synthesizing entirely new forms of contrast. Restoration networks now reclaim dynamic processes from photon-starved data~\cite{morales2022volumetric, cao2024neural, qiao2025neural}; in light-sheet imaging of zebrafish embryos, content-aware models reconstruct crisp subcellular time-lapses in which mitotic spindles can be tracked deep within tissue~\cite{zhang2025deep}. Super-resolution networks extend beyond diffraction, recovering actin filaments at $\sim$120 nm while avoiding the phototoxicity of structured illumination~\cite{wang2019deep, qiao2025fast, jin2020deep, qiao2021evaluation}. Generative models now computationally synthesize biological contrast: in cryo-electron tomograms, self-supervised denoisers reveal membrane–protein complexes otherwise lost in noise~\cite{liu2024j, zhang2024robust, jiang2025cryoccdconditionalcycleconsistentdiffusion}. Together, these advances illustrate a shift: microscopy is evolving into an inference engine where data and algorithms converge to drive discovery~\cite{cunha2024machine, morgado2024rise}.

\begin{figure*}
\centering
\includegraphics[width=0.8\linewidth]{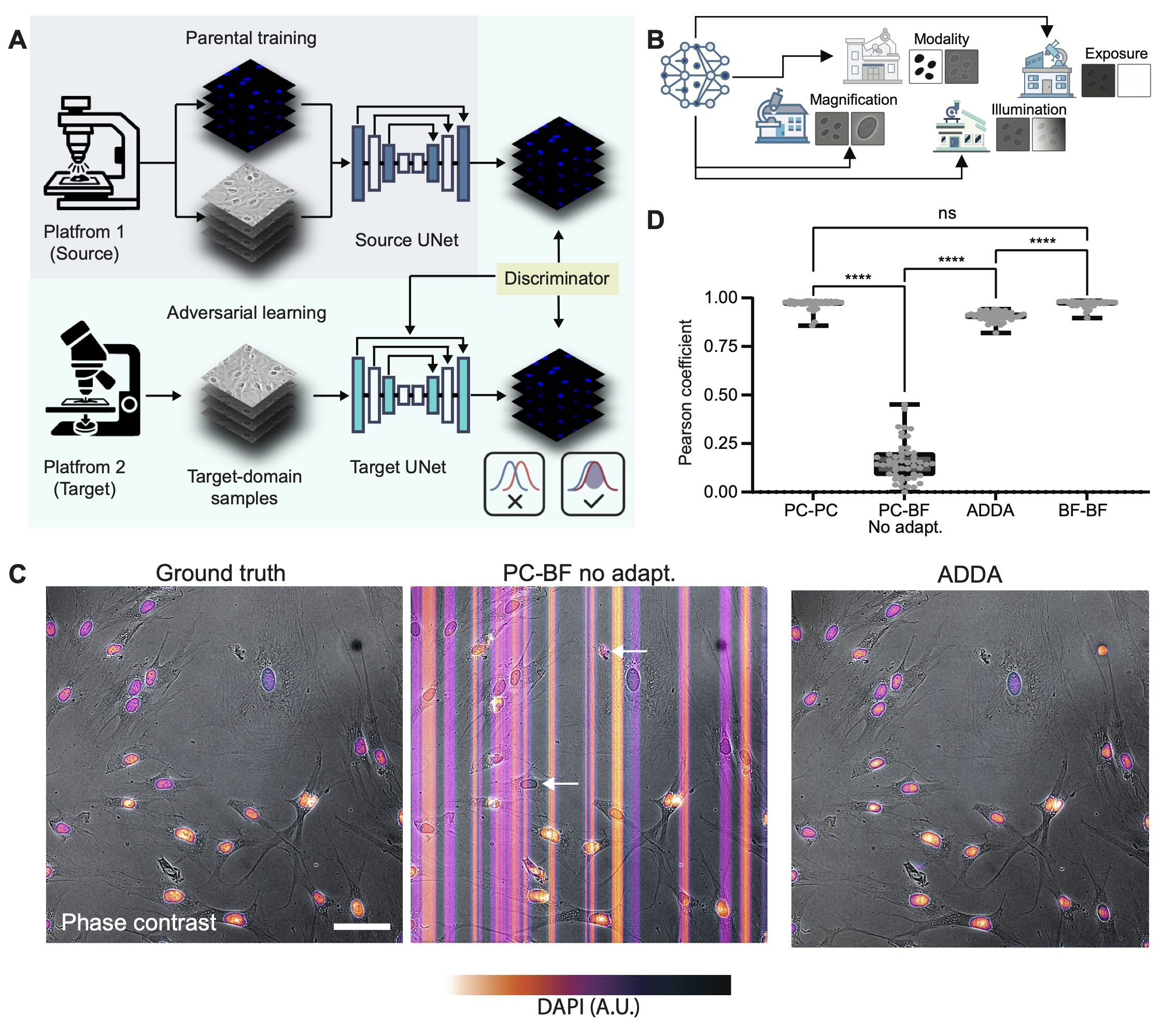}
\caption{
\textbf{Adversarial Discriminative Domain Adaptation (ADDA) enables robust cross-domain image translation in microscopy.} (A) Schematic of the ADDA workflow. Stage 1 (Parental training): train an encoder and predictor (e.g., U-Net) on labeled source images. Stage 2 (Adversarial learning): adapt a target encoder to match the source using unlabeled target data and a discriminator. Stage 3: perform inference on target images with the adapted encoder and fixed predictor. Our variant, SIT-ADDA, selectively fine-tunes a subset of layers to improve robustness across microscopy platforms. (B) This framework generalizes across diverse imaging domain shifts, including differences in magnification, imaging modality, illumination conditions, and exposure settings. (C) Example application to cross-modality nuclear fluorescence prediction. A U-Net trained on phase-contrast data fails to generalize to brightfield inputs (middle), introducing artifacts (white arrows) and missing nuclei. In contrast, ADDA adaptation (right) preserves nuclear morphology and inter-nuclear variability in the target domain. (D) Quantitative evaluation using pixel-wise Pearson correlation shows that direct transfer from phase contrast to brightfield causes accuracy to collapse from $\sim$0.98 to $\sim$0.16, while ADDA adaptation restores performance to 0.91, closely matching source-domain baselines. Scale bar: 100~$\mu$m.
 “ns'' and **** refer to p $\ge$ 0.05 and  <0.0001, respectively.
 }

\label{figure1}
\end{figure*}

Despite these breakthroughs, a persistent barrier limits the broad adoption of AI in microscopy: trained models often fail to generalize across imaging domains~\cite{guan2021domain, yoon2024domain}. Domain shifts, arising from differences in sample preparation, staining protocols, hardware configurations, or acquisition settings, cause substantial changes in image distributions. Even minor shifts make models brittle~\cite{lafarge2017domain, stacke2019closer}, blocking reproducibility and broad use. As such, multiple strategies have attempted to bridge these gaps. Pixel-level style transfer approaches harmonize image appearance across domains~\cite{xing2020bidirectional, chen2019crdoco}, while data-centric methods such as stain normalization and augmentation simulate variability~\cite{mahbod2024improving, gutierrez2022staincut, zhong2025survey}. Recent innovations like ContriMix disentangle biological features from acquisition-specific artifacts to improve training robustness~\cite{nguyen2023contrimix}. Despite tangible progress, these strategies often depend on assumptions about the form of domain variation or require representative labeled data from anticipated use cases~\cite{zhou2022domain}, leaving generalization gaps unresolved~\cite{david2010impossibility, stacke2020measuring}. In this work, we explicitly address a different setting: unsupervised adaptation of a fixed, already-trained predictor to a new acquisition domain using only unlabeled images.

Here, we introduce Subnetwork Image Translation ADDA with automatic depth selection (SIT-ADDA-Auto), a self-configuring scheme for unsupervised adaptation of trained predictors to new acquisition domains. SIT-ADDA-Auto freezes high-level blocks, performs adversarial alignment only in early layers, and uses predictive uncertainty to automatically set adaptation depth on unlabeled target streams. Because our regime is unpaired and label-free, we ground the method in ADDA (Fig.\ref{figure1}A)\cite{tzeng2017adversarial} and compare against full-encoder adaptation~\cite{bhattacharya2024enhancing, pernice2023out, nguyen2023contrimix, guo2025unsupervised, ju2025domain}.

Our design rests on the principle that most microscopy domain shifts perturb low-level optical statistics while leaving the label function (biological semantics) largely stable~\cite{guan2021domain}. This suggests a simple rule: adapt shallow features, freeze deep semantics. We instantiate this rule as a microscopy-optimized ADDA (Fig.\ref{figure1}A)\cite{tzeng2017adversarial}; the uncertainty signal then chooses how far to adapt, avoiding per-site heuristics. Across cross-modality and cross-platform transfers, including low-cost and 3D-printed microscopes (Fig.~\ref{figure1}B), SIT-ADDA-Auto improves reconstruction and downstream performance over conventional ADDA and related methods while preserving high-level semantics under optical shifts. By reducing technical and data demands, SIT-ADDA-Auto offers a lightweight, plug-and-play path to adapt trained models to new microscopes and conditions, broadening access in both advanced facilities and resource-limited settings.

\section*{Results}
\subsection*{Adversarial Domain Adaptation Restores Cross-Modality Image Translation in Microscopy}

To evaluate the impact of domain adaptation on cross-modality translation in microscopy, we first tested whether a U-Net~\cite{ronneberger2015u}, pretrained to predict DAPI fluorescence (nucleus) from phase-contrast images, could generalize to brightfield data (Fig.\ref{figure1}). 
Without adaptation, performance degraded markedly under domain shift, producing stripe artifacts, spurious elongated structures, false-positive nuclei in background regions, and missed detections in dense fields of view (Fig.\ref{figure1}C, middle; white arrows). 
Applying conventional ADDA (Fig.\ref{figure1}A) restored reliable cross-modality translation: most nuclei were accurately delineated, predicted fluorescence intensities closely matched those of the source domain, and both nuclear morphology and inter-nuclear variability were preserved (Fig.~\ref{figure1}C, right). 

To quantify the impact of domain shift on image translation, we computed the pixel-wise Pearson correlation coefficient between predicted and ground-truth images. 
Applying a phase-contrast–trained model directly to brightfield inputs caused the correlation to plummet from $\sim$0.98 to $\sim$0.16, reflecting substantial loss of predictive accuracy.
Following adaptation with conventional ADDA, this performance deficit was largely mitigated: the Pearson coefficient recovered to 0.91 (Fig.~\ref{figure1}D). 
Additionally, Cellpose-based nucleus segmentation of these ADDA predictions showed agreement with ground truth in nuclear centroid positions (Pearson $r \approx 1.0$) and nuclear areas ($r \approx 0.674$ after outlier removal) (Fig. S1).
First, despite lower apparent contrast, brightfield images contain sufficient structure for accurate nuclear labeling.
Second, aligning feature distributions across modalities restores prediction accuracy without requiring target-domain annotations.

\begin{figure*}
\centering
\includegraphics[width=0.85\linewidth]{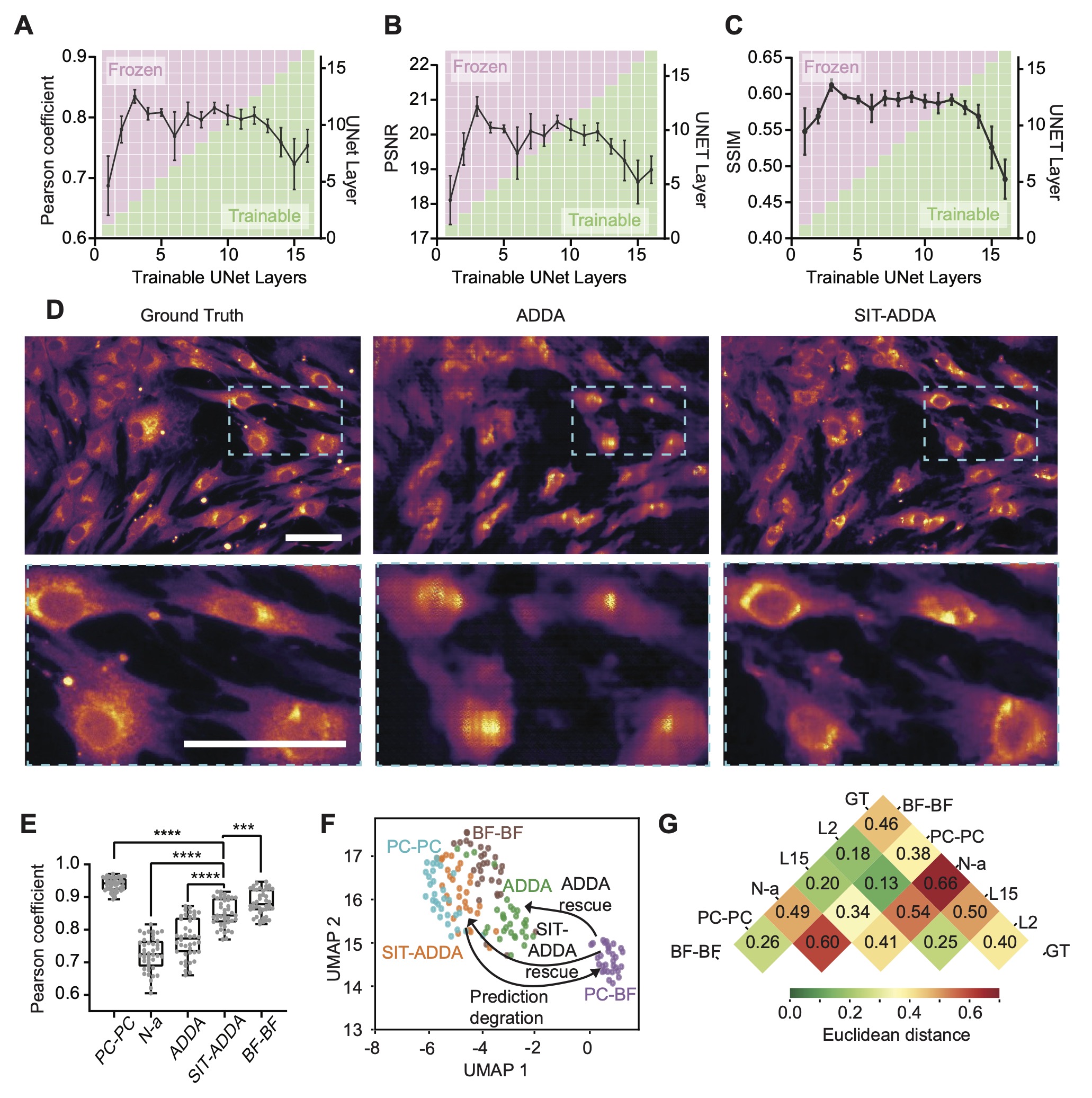}
\caption{\textbf{SIT-ADDA improves cross-modal reconstruction and preserves subcellular detail.} 
(A–C) Ablation results indicate that keeping deeper, semantically meaningful blocks frozen and fine-tuning only the first three convolutional layers (SIT-ADDA) gives the best balance of stability and adaptability — outperforming conventional ADDA (all 16 layers trainable) across Pearson correlation (A), PSNR (B), and SSIM (C).
(D) Representative examples of CD29 fluorescence prediction show that conventional ADDA can introduce cross-shaped and other hallucinated artifacts and fail to recover perinuclear enrichment, whereas SIT-ADDA more faithfully reconstructs perinuclear halo-like signal and preserves peripheral intensity. Scale bars: 100~$\mu$m.
(E) Aggregate prediction accuracy across the test set: conventional ADDA raised Pearson correlation from a no-adaptation (N-a) baseline of 0.73 to 0.77, while SIT-ADDA further increased it to 0.85, reflecting superior recovery of fine-scale subcellular localization. 
(F) Two-dimensional UMAP embeddings of predicted images were computed for four conditions: source-trained (PC-PC and BF-BF), no adaptation (PC-BF), ADDA-adapted (ADDA), and SIT-ADDA-adapted (SIT-ADDA). SIT-ADDA predictions align more closely with the target-domain clusters (PC-PC and BF-BF), indicating improved correction of domain-induced distributional shifts.
(G) Quantification via Euclidean distance between cluster centroids confirms this visual observation: SIT-ADDA predictions are consistently closer to ground-truth and same-modality outputs than those produced by conventional ADDA or by the no-adaptation baseline. *** and **** refer to p <0.001 and <0.0001, respectively.}

\label{figure2}
\end{figure*}

\subsection*{\termlmadda{}: Selective Layer-Freezing Enhances Domain Adaptation in Microscopy}
Conventional ADDA adapts the full encoder but can struggle when domain shifts primarily affect low-level image statistics (illumination, exposure, contrast, sensor noise). Adapting the entire network risks altering higher-level semantic features learned on the source domain and can reduce reconstruction quality. Targeting adaptation to early layers that encode contrast and texture, therefore, may more effectively correct acquisition-specific differences while better preserving domain-invariant representations.

Motivated by this rationale, we introduce Selective Image-Translation ADDA (\termlmadda{}), a transfer-learning–inspired variant that freezes deeper, semantically rich U-Net blocks while adapting the earlier convolutional layers. This reversal of the traditional classification-oriented freezing schedule leverages the empirical fact that deeper layers tend to capture object-level features, whereas early layers encode microscope-specific details~\cite{yosinski2014transferable}. We conducted systematic freezing schedule experiments by testing early layers, deep layers, and dropping individual layers (Figure S2). These results validated our design choice, confirming that the strategy preserves generalizable structural representations while enabling microscope-specific corrections. 

To evaluate the effectiveness of our approach, we tested it on mesenchymal stem cells (MSCs) stained for CD29 (integrin $\beta$1), where perinuclear signal enrichment reflects intracellular trafficking or signaling activity~\cite{de2015integrin, togarrati2018cd}. CD29 produces a spatially subtle, texture-dependent signal that does not coincide with obvious, high-contrast organelles. As a result, faithfully reconstructing CD29 localization requires preservation of fine-scale intensity and structural detail. In contrast, nuclei are easily visualized in transmitted-light microscopy and thus serve as a simpler test case. These features make CD29 a particularly stringent and informative target for evaluating biological validity.

To characterize SIT-ADDA, we conducted an ablation study in which different combinations of encoder and decoder layers were frozen. We evaluated predictions with complementary metrics: Pearson correlation for linear agreement with ground truth, peak signal-to-noise ratio (PSNR) for reconstruction accuracy, and the structural similarity index (SSIM) for perceptual quality and structural integrity. As shown in Figs.~\ref{figure2}A--C, fine-tuning only the first three convolutional layers yielded the best adaptation, improving Pearson correlation (0.84 vs. 0.75), PSNR (20.8 vs. 19.0), and SSIM (0.61 vs. 0.48) compared with ADDA. These improvements across all three metrics demonstrate that \termlmadda{} enhances the quantitative accuracy of reconstructed fluorescence signals.

We also observed that \termlmadda{} outperforms conventional ADDA in capturing fine-scale CD29 localization. While conventional ADDA accurately captured the overall cellular distribution of CD29, it failed to resolve the perinuclear signal enrichment evident in the ground truth, instead introducing cross-shaped artifacts (middle panel in Fig.~\ref{figure2}D). 
In contrast, \termlmadda{} markedly improved subcellular prediction, reconstructing perinuclear halo-like structures with minimal hallucinated features.
It also better preserved signal at the cell periphery, resulting in sharper boundary delineation. 
These improvements further reflect the importance of selectively fine-tuning early layers that govern fine-grained morphological details.

Our quantitative summary of the Pearson coefficient (Fig.~\ref{figure2}E) showed that traditional ADDA improved prediction accuracy over the no-adaptation baseline, raising the correlation from 0.73 to 0.77. \termlmadda{} further increased it to 0.85, representing an overall improvement of ~17\% compared with the baseline. Lastly, we assessed the effectiveness of adaptation strategies by visualizing the distribution of predicted images in a 2D embedding space (Uniform Manifold Approximation and Projection; UMAP), where overlap between the source and target domains reflects successful alignment. As shown in Fig.~\ref{figure2}F, when brightfield input images were processed by a network trained on phase-contrast data, the resulting predictions (PC-BF) exhibited a pronounced shift in distribution, indicating a  degradation in image-to-image translation performance caused by the domain mismatch. 
While traditional ADDA (ADDA) partially corrected this drift, \termlmadda{} effectively aligned the predicted image distribution (SIT-ADDA) with the original target domain cluster. 
Quantitative analysis using Euclidean distance between cluster centroids confirmed that the \termlmadda{} predictions were consistently closer to the ground truth and to direct same-modality predictions (e.g., BF-BF, PC-PC) than those from ADDA or the no-adaptation baseline (Fig.~\ref{figure2}G). This closer correspondence suggests a more successful domain adaptation. In this work, we restrict our comparisons to the conventional ADDA framework, which to the best of our knowledge is the only method that preserves a fixed predictor while adapting only the encoder using unlabeled target-domain data. By contrast, prior approaches that require de novo training of a model on target-domain annotations, or that optimize pixel-level translation objectives (e.g., unpaired style transfer), address fundamentally different problems and therefore do not constitute directly comparable baselines for our formulation. 

\begin{figure*}
\centering
\includegraphics[width=0.95\linewidth]{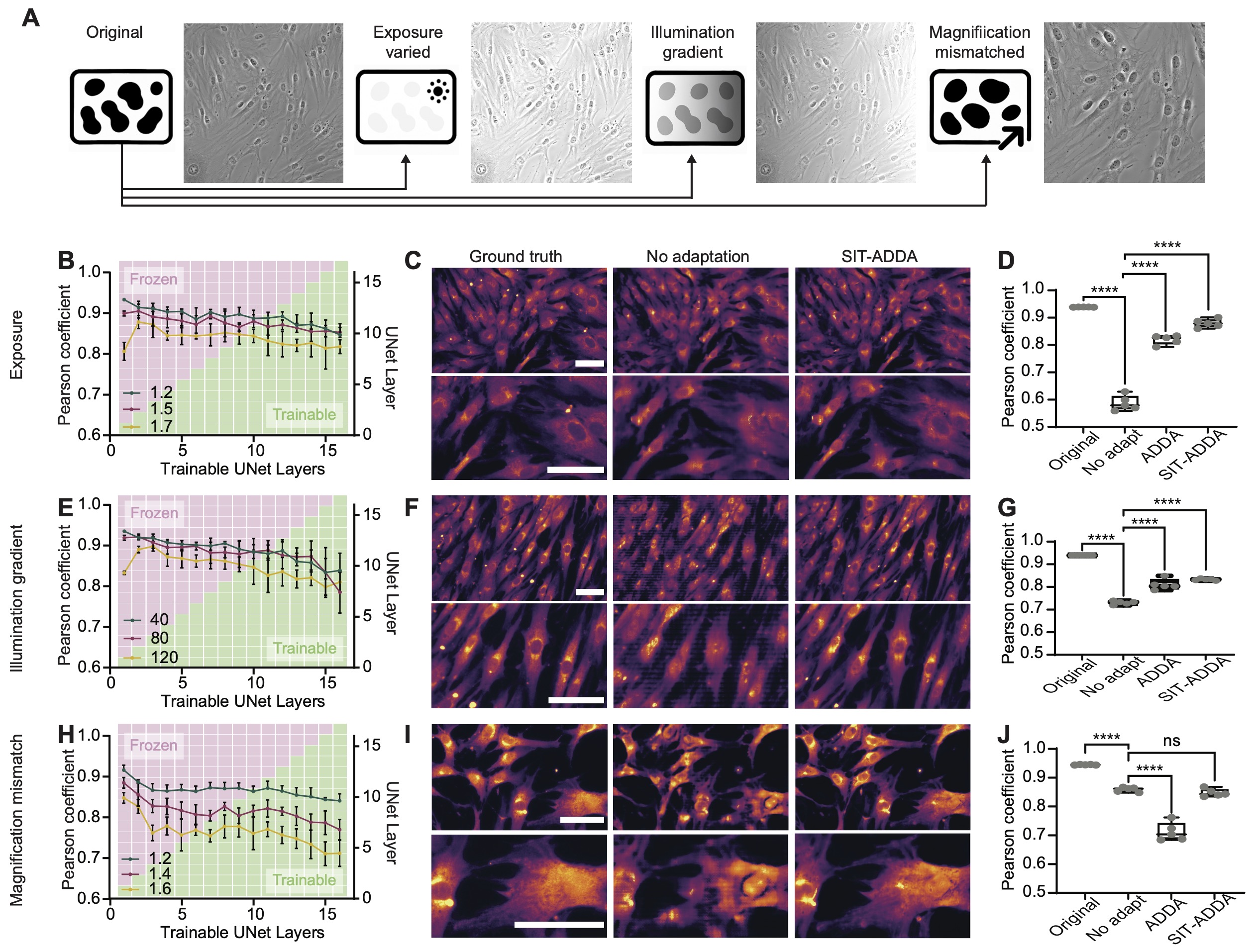}
\caption{\textbf{SIT-ADDA produces prediction improvements across common microscopy domain shifts.} 
(A) Experimental design: we simulated three canonical domain perturbations, namely overexposure, spatial illumination gradient, and magnification (scaling), and evaluated adaptation performance by measuring reconstruction quality (Pearson correlation) across different perturbation strengths.
(B–D) Overexposure results. (B) Summary curves show Pearson correlation as a function of increasing overexposure: SIT-ADDA consistently outperforms conventional ADDA and the no-adaptation baseline. (C) Representative image examples: extreme overexposure that produces saturation and clipping in baseline predictions is corrected by SIT-ADDA, restoring signal delineation and reducing hallucinated features. (D) Quantification: SIT-ADDA increased Pearson correlation by 7.8\% compared to conventional ADDA, representing a statistically significant improvement.
(E–G) Illumination-gradient results. (E) Summary curves for increasing gradient strength indicate greater recovery with SIT-ADDA. (F) SIT-ADDA normalizes spatial bias introduced by misaligned illumination, producing more uniform intensity profiles. (G) SIT-ADDA exceeded conventional ADDA by 13.2\% in Pearson correlation. 
(H–J) Magnification (scaling) results. (H) Pearson versus scaling factor shows that both ADDA and SIT-ADDA achieve smaller gains compared to the other perturbation types. (I) SIT-ADDA can capture global morphology and some subcellular detail (e.g., perinuclear CD29 enrichment) in moderate scaling. (J) Although SIT-ADDA outperforms conventional ADDA by 14.3\%, the source-domain U-Net already shows strong robustness to magnification shifts, yielding stable and reliable predictions. Scale bars: 100~$\mu$m. “ns'' and **** refer to p $\ge$ 0.05 and  <0.0001, respectively.}
\label{figure3}
\end{figure*}

\subsection*{\termlmadda{} Enables Domain Adaptation Across Microscopy Variabilities}
When pre-trained models are applied to new microscopy datasets, domain shifts emerge from variations in illumination, contrast, or modality-specific characteristics.
To systematically evaluate the capacity of \termlmadda{} to generalize across different shifts, we simulated three canonical sources of domain variability (Fig.~\ref{figure3}A): (1) variations in exposure levels, driven by changes in illumination intensity and exposure time (Figs.~\ref{figure3}B--D); (2) spatial illumination gradients introduced by optical misalignment (Figs.~\ref{figure3}E--G); and (3) changes in magnification, reflecting differences in objective lenses and pixel resolution (Figs.~\ref{figure3}H--J);
We synthetically applied these perturbations to controlled image sets, enabling a reproducible assessment of domain adaptation strategies under realistic microscopy conditions. We quantified the extent of these domain shifts by characterizing the perturbed images using cosine similarity, UMAP projections (Fig. S3), and Shannon entropy (Fig. S4). Overexposure and scaling produced the greatest divergence, whereas gradient perturbations caused the least.

We then applied both ADDA and \termlmadda{} to these perturbed datasets, systematically varying the number of frozen layers to identify configurations that best balance generalization and fine-tuning. Across all scenarios, \termlmadda{} outperformed conventional ADDA by roughly 10\% on average relative to the ADDA baseline. As shown in Figs.\ref{figure3} D, G, and J, prediction accuracy improvements were consistent across benchmarking metrics, with gains in Pearson correlation (11.8\% $\pm$ 4.5\% [6.0\%, 19.1\%]), PSNR (14.0\% $\pm$ 3.7\% [8.4\%, 17.7\%]), and SSIM (23.4\% $\pm$ 4.1\% [18.3\%, 32.2\%]). Counterintuitively, fine-tuning only the first layer provided the best adaptation across most tested shifts. In cases of more severe perturbations, such as 1.7-fold overexposure (Fig.\ref{figure3}B, yellow curve) or a 120\% illumination gradient (Fig.\ref{figure3}E, yellow curve), this was not sufficient. In these conditions, incorporating the second layer into training was necessary to achieve robust adaptation. Also, we observed minimal gains from adaptation under magnification-induced shifts (Fig.~\ref{figure3}J), suggesting the baseline U-Net is already relatively scale tolerant. By contrast, both ADDA and SIT-ADDA produced statistically significant gains under overexposure and illumination gradient shifts (Figs.~\ref{figure3}D,\ref{figure3}G).

Closer inspection of the microscopy images revealed in detail how adaptation improved predictions. Overexposed inputs that produced saturation artifacts in baseline predictions were corrected by \termlmadda{}, yielding improved signal delineation (Fig.\ref{figure3}C). Illumination gradients that introduced intensity biases were effectively normalized, producing outputs with uniform intensity profiles (Fig.\ref{figure3}F). In scaled images, \termlmadda{} restored the model’s ability to reconstruct both cellular morphology and perinuclear CD29 enrichment (Fig.\ref{figure3}I). 

To compare AI-based generalization with parametric image restoration, we evaluated histogram normalization, a standard method that mitigates overexposure by rescaling intensities to approximate the original distribution. We found that this method provided only modest improvements in prediction accuracy on overexposed images (Fig. S5). However, it did not correct saturation-induced intensity clipping, resulting in persistent artifacts and performance inferior to ADDA and \termlmadda{}. Together, these results show that selectively fine-tuning early U-Net layers allows \termlmadda{} to outperform parametric methods and robustly accommodate diverse domain shifts.

\begin{figure*}
\centering
\includegraphics[width=0.75\linewidth]{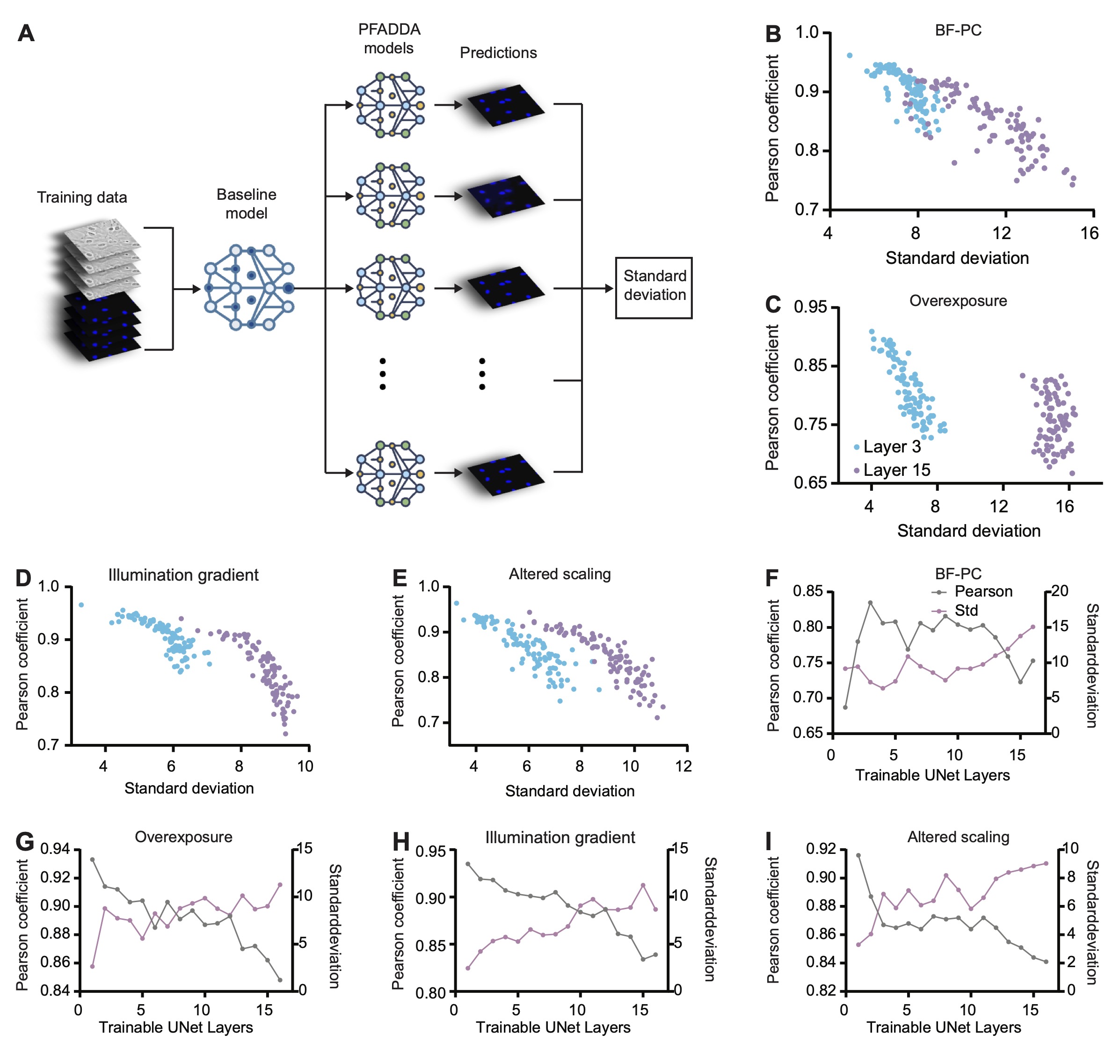}
\caption{\textbf{Ensemble–based uncertainty enables unsupervised optimization of SIT-ADDA.} 
(A) Workflow for estimating epistemic uncertainty. Five independently trained U-Net models were used to form an  ensemble, with per-pixel variance across predictions serving as an approximation of predictive uncertainty. The  mean yields the point estimate, while variance quantifies uncertainty without requiring ground truth. 
(B–E) Across all tested domain shifts, the scatter plots show a distinct separation between methods: SIT-ADDA (Layer 3, blue dots) predictions cluster tightly in the low-variance, high-accuracy region, whereas ADDA predictions (Layer 15, purple dots) spread into a regime of higher variance and lower accuracy.
(F–I) Uncertainty inversely correlates with prediction accuracy. As standard deviation (Std) decreases, Pearson correlation systematically increases, confirming that predictive uncertainty is a reliable surrogate for adaptation performance.}

\label{figure4}
\end{figure*}

\subsection*{Independent Model Ensembles Enable Unsupervised Optimization of \termlmadda{}}
The effectiveness of \termlmadda{} depends on selecting the appropriate number of layers to fine-tune. In practical adaptation scenarios, only unlabeled target-domain images are available, while ground-truth annotations used in our experiments above are not.
To address this challenge, we established an epistemic uncertainty framework that quantifies prediction variability across $K=5$ independently trained U-Net models (Fig.~\ref{figure4}A).
For each input $x$, we calculated the set of predictions ${f_k(x)}_{k=1}^K$; The sample mean of these predictions provided the point estimate, while their sample variance yielded a per-pixel uncertainty score. 
This variance provides a Monte Carlo approximation of the epistemic component of the predictive variance, serving as a  surrogate for full Bayesian integration \cite{kendall2017uncertainties,lakshminarayanan2017simple}. Variability across independently trained models has been shown to yield well-calibrated uncertainty estimates, to track true error in dense prediction tasks such as optical flow and medical image segmentation \cite{ilg2018uncertainty,park2024probabilistic,imboden2023trustworthy}, and to detect distributional shifts \cite{ovadia2019can}. Moreover, because independent stochastic gradient descent runs converge to different connected minima, their predictions approximate posterior samples \cite{garipov2018loss}.

To illustrate how our uncertainty framework guides adaptation, we first focused on the cross-modal scenario (Fig.~\ref{figure2}). In this setting, fine-tuning three layers with SIT-ADDA had previously outperformed full-model adaptation with fifteen trainable layers in traditional ADDA. This case provided a benchmark for evaluating whether the distribution of standard deviation (Std) values, computed for individual inputs, could reliably capture differences in model performance (Fig.\ref{figure4}B). The analysis revealed a clear separation between the two conditions: SIT-ADDA exhibited consistently lower Std values and higher Pearson correlation coefficients relative to ADDA. 
This pattern was robust across  domain shifts induced by overexposure (Fig.\ref{figure4}C), illumination gradients (Fig.\ref{figure4}D), and scaling perturbations (Fig.\ref{figure4}E). To further examine the link between uncertainty and predictive accuracy, we compared Std values directly with Pearson correlation coefficients (Figs.\ref{figure4}F--I), revealing an inverse relationship in which lower Std values coincided with higher Pearson correlations. Accordingly, the ensemble-derived standard deviation provides a practical criterion for selecting adaptation depth, enabling unsupervised optimization of \termlmadda{} across the tested domain shifts.

\begin{figure*}
\centering
\includegraphics[width=0.8\linewidth]{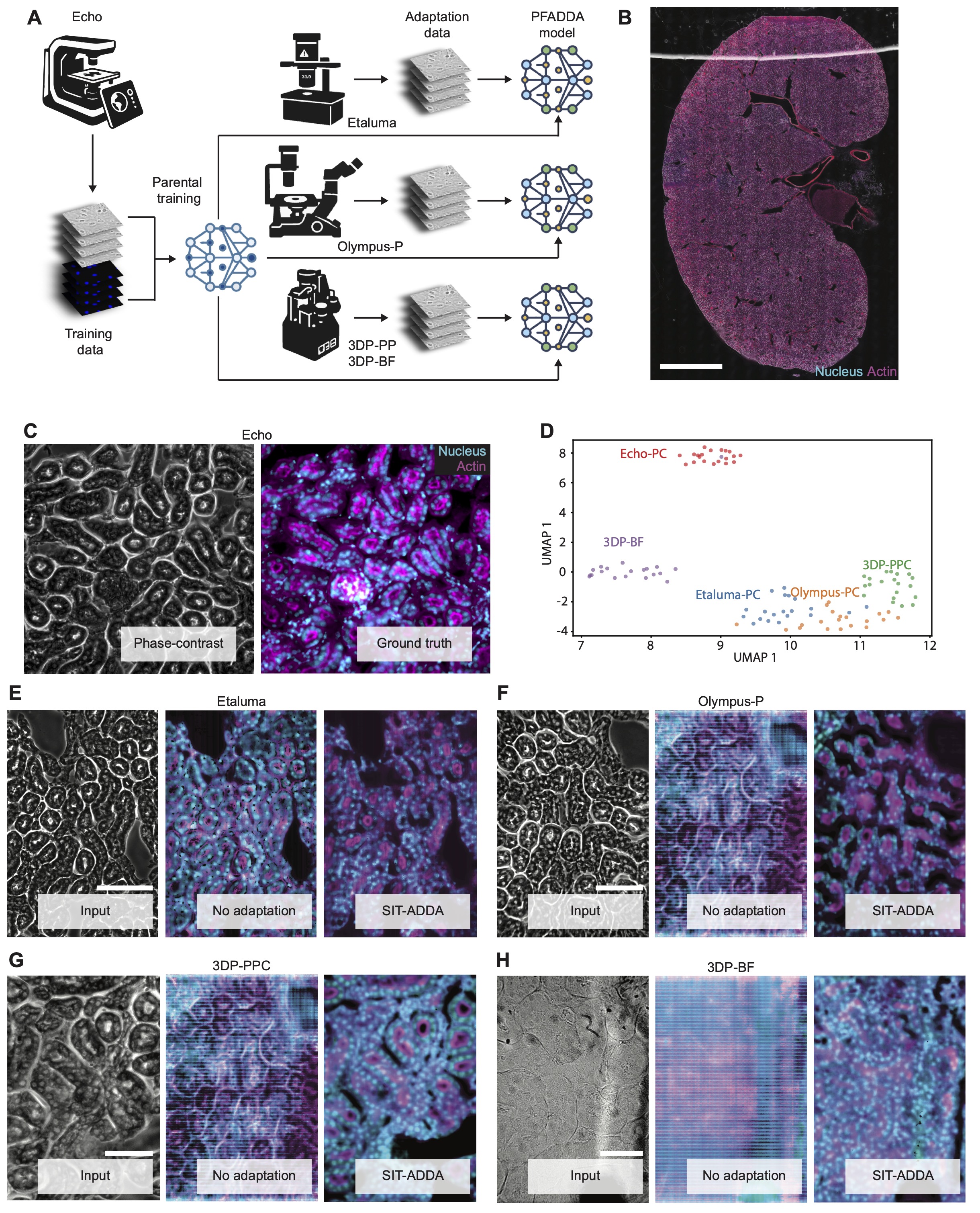}
\caption{\textbf{SIT-ADDA enables robust model transfer across diverse microscopy systems.} 
(A) Experimental setup: models trained on Echo microscope data were tested on four distinct target platforms spanning a spectrum of optical performance, from lab-grade instruments to citizen-science tools, to evaluate the ability of SIT-ADDA to recover fluorescence-like predictions. 
(B) Mouse kidney tissue labeled with DAPI (nucleus, cyan) and phalloidin (actin, magenta) was used to assess adaptation performance. Scale bar: 500~$\mu$m.
(C) Source domain reference: predictions from the Echo-trained model closely matched ground truth, accurately reconstructing actin filaments and nuclear morphology. 
(D) Feature-space analysis revealed five distinct and well-separated clusters, corresponding to the Echo source images and the four target platforms, validating pronounced distributional shifts.
(E–H) Representative inputs and predictions for each target platform. Without adaptation, predictions failed to capture meaningful nuclear or cytoskeletal structure. In contrast, SIT-ADDA consistently restored biologically plausible features. Scale bars: 100~$\mu$m.}

\label{figure5}
\end{figure*}

\begin{figure*}
\centering
\includegraphics[width=0.9\linewidth]{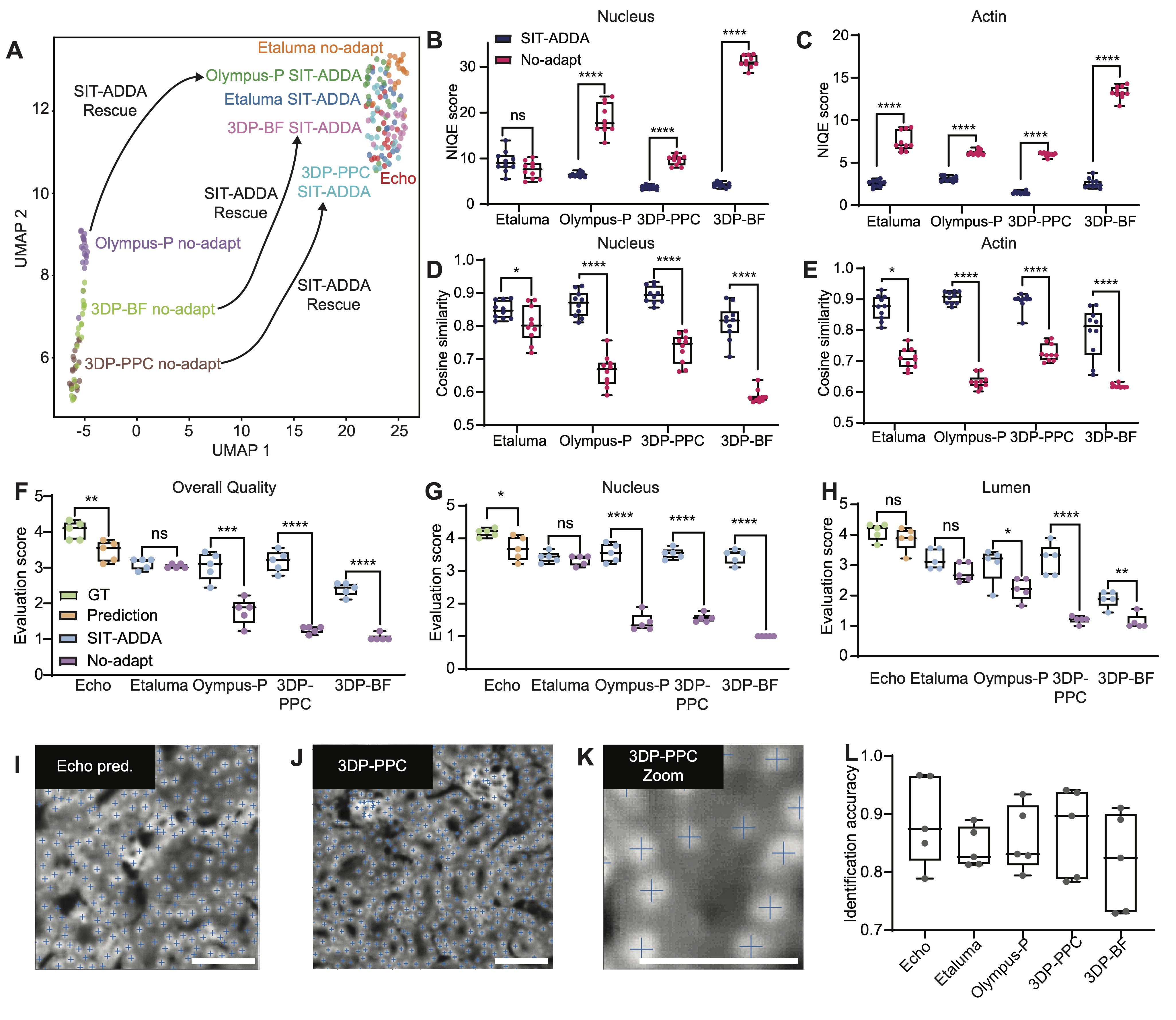}
\caption{\textbf{SIT-ADDA improves cross-platform prediction quality and preserves downstream analytical utility.} 
(A) UMAP embedding of predicted images demonstrates that SIT-ADDA shifts the non-adapted outputs toward the original Echo prediction cluster, indicating effective cross-domain alignment. Here, Echo predictions are used as a platform-consistent reference because co-registered fluorescence is unavailable on target devices.
(B–C) NIQE scores, where lower values correspond to sharper and more natural images, demonstrate that SIT-ADDA consistently improves perceptual quality for both DAPI (nucleus, B) and phalloidin (actin, C) predictions relative to no adaptation. 
(D–E) Cosine similarity analysis further confirms that adapted predictions are more closely aligned with Echo references than their non-adapted counterparts, for both DAPI (nucleus, D) and phalloidin (actin, E). 
(F–H) Results from a blinded expert questionnaire show that SIT-ADDA-adapted images received higher ratings for (F) overall image quality, (G) nuclear clarity, and (H) lumen clarity. Adapted Etaluma predictions approached the quality of Echo outputs, while even lower-grade systems such as 3DP-BF benefited in terms of nuclear identification (G). 
(I–L) Cellpose-based segmentation analysis reveals that adapted predictions achieve nucleus identification accuracy comparable to Echo outputs, with near 90\% centroid detection accuracy across all platforms. Scale bars: 50~$\mu$m. `ns', *, **, ***, and **** refer to p $\ge$ 0.05, <0.05, <0.01, <0.001, and <0.0001, respectively.}
\label{figure6}
\end{figure*}

\subsection*{\termlmadda{} Enables Robust Model Transfer Among Diverse Imaging Platforms}

To evaluate real-world utility, we next tested whether \termlmadda{} can support cross-platform adaptation, a challenging scenario where multiple domain shifts arise simultaneously (Fig.~\ref{figure5}A). To this end, we employed mouse kidney tissue sections (Fig.~\ref{figure5}B) labeled with DAPI (nucleus) and phalloidin (F-actin) stains. 
As depicted in Fig.~\ref{figure5}A, training data for the baseline models were acquired on a high-end epifluorescence imaging system (Echo). 
For cross-platform testing, we introduced images from four microscopes representing a broad spectrum of optical performance: laboratory-grade optics (Etaluma), educational-grade systems (Olympus paired with an iPhone, as used in K–12 environments; Olympus-P), advanced hobbyist platforms (OpenFlexure 3D-printed microscope with pseudo-phase contrast modifications; 3DP-PPC), and citizen-science tools (OpenFlexure in a standard configuration; 3DP-BF). See Fig. S6 for 3D-printed microscope configurations. This experimental design also demonstrates a unique application of \termlmadda{}, enabling compensation for the absence of fluorescence imaging capabilities in the latter three platforms. Representative images from the five microscopy systems and their quantitative characterizations are found in Figs. S7 and S8. Before testing domain shifts, we verified that the Echo dataset enabled accurate U-Net predictions of actin and nuclei (Fig.~\ref{figure5}C). Actin predictions preserved filamentous structures with luminal enrichment, while DAPI predictions matched nuclear morphology and distribution. These Echo-trained predictions serve as a stable reference for all downstream assessments, allowing paired comparisons of adapted versus non-adapted outputs per field of view.

Across all four tested target domains, which formed distinct clusters in feature space (Fig.\ref{figure5}D), \termlmadda{} substantially improved virtual staining relative to the no-adaptation baseline (Figs.\ref{figure5}E--H). This effect was consistent across microscopes of markedly different optical quality, indicating that the gains are not idiosyncratic to a single device but reflect robustness to the combined shifts of contrast, illumination and aberrations encountered in routine imaging. On the Etaluma (Fig.\ref{figure5}E) and Olympus-P (Fig.\ref{figure5}F) datasets in particular, \termlmadda{} produced sharper nuclear and cytoskeletal reconstructions with reduced background artifacts, whereas baseline models yielded distorted outputs lacking biologically meaningful features. In the more challenging OpenFlexure domains, both 3DP-PPC ((Fig.\ref{figure5}G) and 3DP-BF (Fig.\ref{figure5}H) inputs presented significant obstacles to accurate prediction, owing to their inherently low contrast, uneven illumination, and pronounced optical aberrations. Nevertheless, \termlmadda{} successfully adapted the model to recover actin-rich cell boundaries and discrete nuclear features that were barely discernible in the raw inputs. Notably, improvement on the lowest-grade platform (3DP-BF) was most evident for nuclei, consistent with the method recovering features that are reliably present despite degraded inputs.

Using UMAP (Fig.\ref{figure6}A), we verified that \termlmadda{} shifts unadapted images closer to the Echo prediction cluster across all microscopy platforms. To assess adaptation effectiveness without ground truth, we used three complementary metrics: (1) image quality and cosine similarity to Echo predictions, measuring resemblance to reference images; (2) expert evaluation, benchmarking biological realism and interpretability; and (3) nucleus segmentation performance, testing utility for downstream analysis. These readouts were pre-specified to reflect the biology in our samples. Nucleus-centric endpoints capture cell counts and enable density and morphology analyses, while lumen clarity reflects essential renal tubule architecture characteristics. On lower-grade platforms without target-domain fluorescence ground truth, this combination provides orthogonal evidence: perceptual quality, blinded biological interpretability, and quantitative downstream performance. Also, because per-field fluorescence cannot be acquired on these devices, direct prediction-to-ground-truth comparisons are not possible. Our evaluation instead separates a distribution-alignment proxy (similarity to Echo) from independent endpoints not anchored to Echo (expert judgment and Cellpose accuracy). Consistent gains across these views would indicate that SIT-ADDA improves biologically meaningful signal rather than fitting only to Echo-specific appearance.

We first evaluated image quality using the Naturalness Image Quality Evaluator (NIQE), where lower scores indicate sharper structures and reduced distortion. \termlmadda{} consistently lowered NIQE for both DAPI (Fig.\ref{figure6}B) and actin predictions (Fig.\ref{figure6}C). Adapted images also exhibited higher cosine similarity to the original predictions (Fig.~\ref{figure6}D–E). Finally, Laplacian variance, which reflects high-frequency artifacts, was reduced, while root mean square contrast increased, indicating improved structural clarity (Fig. S9). For each platform we performed paired comparisons across matched fields of view between adapted and non-adapted outputs; significance indicators in Figs. 6B--E reflect two-sided nonparametric paired tests with multiplicity control across metrics and platforms, with exact procedures detailed in Methods.

Beyond quantitative image quality metrics, we assessed biological interpretability through expert evaluation. A blinded questionnaire (example shown in Fig. S10) of nine experienced microscopists asked participants to rate overall image quality (Fig.\ref{figure6}F), clarity of nuclear visualization (Fig.\ref{figure6}G), and definition of luminal structures (Fig.\ref{figure6}H). Adapted images consistently received higher scores across all three metrics (Fig. S11). Specifically, \termlmadda{} successfully adapted inputs from Etaluma, Olympus-P, and 3DP-PPC, yielding overall image quality and nuclear and lumen interpretability comparable to Echo predictions. This pattern held for both high- and low-grade instruments, reinforcing that the perceived gains are not device-specific but reflect improvements relevant to end-users’ qualitative judgments. 3DP-BF adaptations showed improvement primarily in nuclear identification (Fig.\ref{figure6}G) but not in lumen morphology (Fig.\ref{figure6}H), likely reflecting the substantially lower quality of the raw inputs. The consistent improvement in nuclear identification likely reflects the visibility of nuclei across imaging conditions, enabling robust recovery by \termlmadda{}.

Lastly, we assessed a critical utility of \termlmadda{}-recovered images by performing a standard downstream analysis with Cellpose nucleus segmentation (Fig.\ref{figure6}I–K). We focused on nucleus centroid labeling (blue crosses), a key metric for cell counting and subsequent packing or morphological analyses. We prioritize centroid detection because it is the decision-making endpoint in regular cell-based assays (confluence estimation, regional cell counts) and is less confounded by ambiguous nuclear boundaries across devices. From our analysis, labeling accuracy, defined as the fraction of true positives among all labels, showed no significant differences between Echo predictions and adapted conditions, with all reaching nearly 90\% accuracy (Fig.\ref{figure6}L). Across n = 10 manually annotated fields per platform, adapted predictions achieved nucleus centroid detection comparable to Echo references, indicating that perceived improvements translate into routine quantitative workflows. Visual inspection confirmed that the centroids identified by Cellpose coincided with the corresponding intensity centers (Fig.\ref{figure6}K). Together, these results show that \termlmadda{} extends deep learning–based virtual staining across diverse imaging platforms, from commercial microscopes to fully 3D-printed systems, while maintaining the accuracy needed for reliable biological interpretation. It enables robust adaptation from high-end fluorescence instruments to resource-limited settings and provides a practical framework for making advanced virtual staining accessible in both scientific and educational contexts.

\section*{Discussion}
Our results indicate that full-network adaptation is often unnecessary and present a simple, stable recipe for handling domain shifts in microscopy. Conceptually, our finding is consistent with transfer-learning observations that early layers encode task-agnostic, low-level statistics, whereas deeper layers capture task-specific semantic structure that should be preserved~\cite{yosinski2014transferable, wang2025transfer, kornblith2019better}. In our setting, constraining updates to early filters absorbs acquisition variability (contrast, illumination, noise, optics) without disturbing downstream semantic representations of subcellular morphology. A mechanistic next step is to quantify layer-wise drift between source and target using centered-kernel alignment or related similarity measures to test this hypothesis directly~\cite{klabunde2025similarity, kornblith2019similarity}.

Operationally, we show that predictive uncertainty can act as a principled, label-free signal to choose adaptation depth in real deployments. While uncertainty has most often been used to prioritize labeling or flag low-confidence inputs, recent work has leveraged it for model optimization and hyperparameter selection~\cite{liu2024uq, wenzel2020hyperparameter}. Building on that perspective, we repurpose prediction variance as a configuration selector for how many early layers to fine-tune, providing an actionable control knob when paired targets are unavailable. This converts a potentially brittle design choice into a data-dependent decision, aiding non-expert users facing new microscopes or acquisition settings.

SIT-ADDA does have limitations that point to immediate refinements. First, magnification (scale) shifts remain challenging: the non-adapted model shows partial robustness, and early-layer adaptation alone does not reliably close the gap. In practice, this can be mitigated with scale-aware modules (e.g., spatial transformers~\cite{jaderberg2015spatial} or multi-scale front ends such as feature-pyramid networks~\cite{lin2017feature}) and simple pre-deployment diagnostics (e.g., resampling target images to match source pixel size) to distinguish cases where preprocessing rather than adversarial alignment is appropriate~\cite{kornblith2019similarity}. Second, our formulation implicitly assumes the target prediction distribution resembles the source (e.g., similar specimens). This can be relaxed by (i) broadening pretraining with diverse, multi-source microscopy and self-supervised contrastive objectives to bias encoders toward content-centric features~\cite{huang2023self, azizi2021big}, and (ii) incorporating test-time, self-supervised adjustment that updates only normalization statistics or lightweight adapters via invariance objectives, stabilizing predictions without forcing source-like outputs~\cite{sun2020test, wang2020tent, karani2021test}.

In summary, by freezing high-level semantics and adapting low-level, domain-variant filters, SIT-ADDA offers a robust, easily deployed solution for virtual-staining domain shifts. Across modality, exposure, and illumination changes it outperforms conventional ADDA. The accompanying unsupervised ensemble procedure uses predictive variance to select the number of trainable layers, enabling no-ground-truth deployment. These properties make SIT-ADDA well suited for cross-platform transfer, including to compact 3D-printed microscopes lacking fluorescence, and offer a practical template for domain adaptation in scientific imaging. Beyond microscopy, the idea of subnetwork-level adaptation informed by uncertainty may translate to medical imaging, materials characterization, and remote sensing, where labeled targets are scarce and distribution shifts are common.

\section*{Methods}

\subsection*{Microscopy Datasets and Imaging Protocols}
Human bone marrow-derived MSCs (ATCC, PCS-500-012) were cultured following the manufacturer’s instructions and established protocols\cite{imboden2021investigating, imboden2023trustworthy}. For immunofluorescence, cells were washed with PBS+/+ and fixed in 4\% paraformaldehyde (Thermo Fisher Scientific, 28908) in PBS+/+ (Gibco), followed by PBS+/+ washes. Blocking was performed with 2\% donkey serum (Sigma-Aldrich, D9663-10ML) and 0.5\% Triton X-100 (Sigma-Aldrich, T8787-50ML). After two PBS+/+ washes, cells were incubated with primary antibody solution (0.5\% BSA, 0.25\% Triton X-100, and primary antibody) for 30 min and washed twice with PBS+/+. Secondary antibody staining was performed for 30 min, followed by two PBS+/+ washes. Imaging was conducted using phase-contrast, brightfield, and fluorescence microscopy (Etaluma LS720, Lumaview 720/600 software) with a 20$\times$/0.45NA phase-contrast objective (Olympus, LCACHN 20XIPC).

For the mouse kidney dataset, a pre-labeled cryosection (FluoCells$^{\mathrm{TM}}$ Prepared Slide~\#3, Invitrogen, F24630; 16~$\mu$m) stained with Alexa Fluor 568 phalloidin and DAPI was imaged. Fluorescence images were acquired in the DAPI and Alexa Fluor 568 channels using both an Echo Revolve microscope and an Etaluma LS720 microscope equipped with a 20$\times$/0.45NA phase-contrast objective (Olympus, LCACHN 20XIPC). Transmitted light images were additionally acquired using an Olympus CKX41 microscope with a Plan N 20$\times$ Ph1 phase-contrast objective coupled to an iPhone 4 via the eyepiece, and a 3D-printed OpenFlexure microscope fitted with an Olympus PlanC N 40$\times$ Ph2 phase-contrast objective. 
The illumination module of the OpenFlexure system was modified with a 30-mm LED ring to produce pseudo–phase-contrast images.

\subsection*{Image Translation Model Architecture}
The generative model employed in this work is based on a U-Net architecture with eight levels of downsampling and upsampling, designed for image-to-image translation tasks at a resolution of $1056 \times 1056$ pixels. 
The encoder and decoder components consist of convolutional and transposed convolutional layers, respectively, both utilizing $4 \times 4$ kernels with a stride of 2 and padding of 1 to appropriately preserve spatial dimensions during downsampling and upsampling operations. 
Skip connections are incorporated between corresponding encoder and decoder layers to facilitate gradient flow and retain fine-grained spatial information throughout the network. 
Normalization layers are applied after each convolutional layer except the outermost output layer. 
Activation functions include Leaky ReLU in the encoder and ReLU in the decoder, with a final Tanh activation to produce output images normalized between $-1$ and $1$.

The discriminator model follows a PatchGAN design, structured as a three-layer convolutional neural network. 
Each convolution uses $4 \times 4$ kernels with a stride of 2 and padding of 1, progressively increasing the number of feature channels while reducing spatial resolution. 
Leaky ReLU activations and normalization layers are employed after each convolution except the last, which outputs a single-channel spatial map indicating the discriminator’s real/fake classification confidence for each image patch. 
This patch-level discrimination allows the model to focus on local structure, encouraging high-frequency detail consistency in the generated images.

\subsection*{Adaptation Model Implementation}

\paragraph{Traditional ADDA.}
We implemented Adversarial Discriminative Domain Adaptation (ADDA) to enable unsupervised domain adaptation. 
Let $\mathcal{D}_S = \{(x_S, y_S)\}$ denote the labeled source domain, and $\mathcal{D}_T = \{x_T\}$ the unlabeled target domain.
The goal of the adaptation is to learn a target feature representation that aligns with the source feature space such that a classifier trained on $\mathcal{D}_S$ generalizes effectively to $\mathcal{D}_T$.

\textbf{Step 1: Source model training.} 
A source encoder $F_{\theta_S}: \mathbb{R}^{H \times W \times C} \rightarrow \mathbb{R}^{H \times W \times C}$ is trained using the mean squared error (MSE) loss:
$\displaystyle \mathcal{L}_{\text{MSE}}(y, \hat{y}) = \mathbb{E} \left\| y - \hat{y} \right\|_2^2$,
where $y$ and $\hat{y}$ denote the ground truth and predicted pixel values, respectively:
\[
\min_{\theta_S} \ \mathcal{L}_{\text{MSE}}(F_{\theta_S}(x_S), y_S).
\]

\textbf{Step 2: Adversarial Discriminative Domain Adaptation.}  
Let $F_{\theta_S}$ denote the source encoder, $F_{\theta_T}$ the target encoder with parameters $\theta_T$, and $D_\phi$ the discriminator with parameters $\phi$. The discriminator maps feature representations to a binary domain label, $D_\phi: \mathbb{R}^{H \times W \times C} \rightarrow \{0,1\}$.  

The target encoder and discriminator are jointly trained by iteratively optimizing discriminative and encoding losses in an adversarial manner. Specifically, the discriminator is trained to classify whether the features originate from the source or target domain, while the encoder is trained to fool the discriminator:
\[
 \min_{\theta_T} \max_{\phi} \ 
 \mathbb{E}_{x_S}[\log D_\phi(F_{\theta_S}(x_S))] \;+\; 
 \mathbb{E}_{x_T}[\log (1 - D_\phi(F_{\theta_T}(x_T)))].
\]

The source encoder $F_{\theta_S}$ is frozen during adaptation, and the target encoder $F_{\theta_T}$ is initialized from $F_{\theta_S}$. During the iterative optimization process, $D_\phi$ is frozen while $F_{\theta_T}$ is optimized to align feature distributions, and $F_{\theta_T}$ is frozen while $D_\phi$ is optimized to discriminate between the outputs from the source and target encoders.

\paragraph{Subnetwork Image Translation ADDA (\termlmadda{}).}
Due to the limited amount of image data, properly limiting the model class size is essential to achieve proper generalization.
To this end, we explored freezing the model weights of some layers in the target encoder while performing domain adaptation. Let $F_T = [f_1, f_2, ..., f_n]$ denote the encoder's $n$ convolutional layers. 
To determine the most effective weight-freezing strategy for domain adaptation in microscopy, we empirically evaluated three different configurations: (1) freezing the first \(k\) layers while training $f_{k+1}$ to $f_n$, (2) freezing the last \(k\) layers while updating the earlier $f_{1}$ to $f_{k-1}$ ones, and (3) freezing all layers while only training $f_k$.

In doing this, we adopted a selective freezing scheme in \termlmadda{}, where only the first \(k\) convolutional blocks of the target encoder/decoder $F_{\theta_T}$ are trainable, and the remaining blocks are frozen:
$\displaystyle \theta_{f_1}, ..., \theta_{f_k} \text{ are updated}; \quad \theta_{f_{k+1}}, ..., \theta_{f_n} \text{ are fixed}.$
We empirically tested values of $k \in \{1, ..., 16\}$ and found that tuning just the first 1--3 blocks (\(k \in \{1, 2, 3\}\)) consistently yielded the most robust adaptation performance across diverse domain shifts in the images tested (see Figs.~\ref{figure2}A--C,~\ref{figure3}B,E,H).

Baseline rationale. For fairness, we benchmarked only against conventional ADDA, which adheres to the same unsupervised adaptation setting (fixed source-trained predictor; unlabeled target images). Methods that require target labels, retrain the predictor from scratch, or optimize image realism objectives address different problem formulations and would confound comparisons. We therefore restrict baselines to full-encoder ADDA and a strong parametric restoration method, isolating the effect of subnetwork freezing within the ADDA framework.

\subsection*{Image Data Perturbation}
To systematically evaluate the robustness of our model under acquisition variability, we synthetically perturbed MSC CD29 images using three canonical transformations: scaling, overexposure, and illumination gradient. 
Image scaling was performed by simulating a digital zoom effect. 
Each image was cropped at the center by a factor inversely proportional to the zoom factor (x1.2, x1.4, and x1.6), followed by resizing to the original resolution. This mimics optical zoom or focal shift artifacts that may alter cellular context or spatial resolution. 
Overexposure was simulated by increasing the image brightness using a linear multiplicative factor (x1.2, x1.5, and x1.7) applied to pixel intensities via brightness enhancement. 
This perturbation emulates sensor saturation or improper illumination calibration often encountered in high-throughput imaging setups.
Illumination gradient was introduced by superimposing a left-to-right horizontal intensity gradient across the image. 
The gradient values linearly increased from 0 to a maximum value (40, 80, or 120), then added channel-wise to the original image before clipping to the valid dynamic range (0–255). 
This simulates uneven illumination across the field of view due to shading or vignetting artifacts in microscopy.
Each perturbation was applied independently to the full set of CD29 images, resulting in a set of systematically distorted inputs used to assess domain generalization performance.

\subsection*{Training \& Evaluation Setup}
Each dataset was split into training, validation, and test subsets (7:1.5:1.5). Models were implemented in PyTorch~\cite{paszke2019pytorch}, with images resized to 1024×1024 pixels and processed in batches of 8. A U-Net backbone was pretrained on the source domain, after which selected convolutional blocks were frozen to various depths and adapted using our \termlmadda{} approach with domain-adversarial training. Models were trained for up to 200 epochs with the Adam optimizer, using a linear learning-rate decay and checkpointing based on validation Pearson correlation. Grid sweeps were performed across discriminator learning rates ($10^{-3}$–$10^{-6}$) and 16 weight-freezing configurations. For evaluation, predictions were compared against ground truth using Pearson correlation, PSNR, and SSIM; predictions were histogram-matched before computing PSNR and SSIM. For each configuration, five independent models were trained, and performance was reported as the mean and standard deviation across runs, with unstable or non-convergent models excluded.
For unsupervised optimization of \termlmadda{}, five models were trained per configuration, and pixel-wise prediction standard deviations were averaged to quantify uncertainty. This was validated on cross-platform and domain-shifted CD29 datasets, where higher prediction accuracy consistently corresponded to lower predictive uncertainty. The optimal learning rate and freezing configuration were therefore selected as those minimizing uncertainty.

\subsection*{Metrics of Prediction Accuracy Evaluation}
To evaluate how the predicted images $\hat{y}$ match the ground truth $y$, we used three complementary metrics.
We quantify pixel-wise correlation using the Pearson correlation coefficient, which measures global linear intensity agreement between the reference image \(y\) and the prediction \(\hat y\): \(\rho(y,\hat y)=\frac{\sum_{p\in\Omega}(y_p-\bar y)(\hat y_p-\overline{\hat y})}{\sqrt{\sum_{p\in\Omega}(y_p-\bar y)^2}\,\sqrt{\sum_{p\in\Omega}(\hat y_p-\overline{\hat y})^2}}\). Here, \(y_p\) and \(\hat y_p\) are the intensities at pixel \(p\in\Omega\), \(\Omega\) is the set of all pixels with cardinality \(|\Omega|\), \(\bar y=\frac{1}{|\Omega|}\sum_{p\in\Omega}y_p\) and \(\overline{\hat y}=\frac{1}{|\Omega|}\sum_{p\in\Omega}\hat y_p\) are the corresponding sample means. 
Second, the peak signal-to-noise ratio (PSNR)
$\mathrm{PSNR}(y,\hat y) = 10 \log_{10}!\left(\frac{L^2}{(y - \hat y)^2}\right)$,
where $L$ denotes the maximum pixel value, quantifying reconstruction quality.
Third, the structural similarity index (SSIM)
$\mathrm{SSIM}(y,\hat y) = \frac{(2\mu_y\mu_{\hat y}+c_1)(2\sigma_{y\hat y}+c_2)}{(\mu_y^2+\mu_{\hat y}^2+c_1)(\sigma_y^2+\sigma_{\hat y}^2+c_2)}$,
with $\mu$, $\sigma$, and $\sigma_{y\hat y}$ representing patch-wise means, variances, and cross-covariances, and $c_1,c_2$ stabilization constants, capturing structural and perceptual consistency.

\subsection*{UMAP Visualization, Quality Metrics, and Segmentation Evaluation}
To generate UMAPs for visualizing the image distribution, each image was embedded using the pretrained BiomedCLIP~\cite{zhang2023biomedclip} image encoder to produce 512‑D feature vectors. 
Pairwise Euclidean distances between these vectors were computed and supplied as the metric for UMAP via the umap‑learn v0.5.3 implementation~\cite{mcinnes2018umap-software}. 
UMAP was configured to project into two dimensions with $n_{neighbors}$ = 15, $min_{dist}$ = 0.1, and metric = ``euclidean'', with all other parameters kept at their defaults.
The resulting 2D embedding was visualized as a scatter plot to illustrate the distribution of images in feature space.

NIQE scores were computed by first fitting a custom model to the Echo prediction images as the reference set using MATLAB’s \texttt{fitniqe} function. 
Following model fitting, each test image was similarly preprocessed and the NIQE score was calculated using the \texttt{niqe} function together with the fitted model parameters. 
Finally, per‐image NIQE values were aggregated by calculating the mean score for each test set, yielding a comparison of image quality metrics across all experimental conditions.

Image similarity was quantified by extracting a 512-D feature vector from ResNet-18’s global average-pooled convolutional output. Each vector was L2-normalized and the ResNet-18 model was pretrained on ImageNet (ILSVRC2012). Images were resized to 224$\times$224 px and grayscale images were replicated across the three input channels to match the network’s RGB input. Cosine similarities between each test vector and all reference embeddings were then computed. For each image, the maximum similarity (clamped to [–1,+1]) was retained, and these maxima were then averaged across samples to yield the mean similarity score per condition

Nuclear segmentation was carried out in MATLAB (R2024a; The MathWorks) using the Cellpose algorithm (v1.0)~\cite{stringer2021cellpose}. 
Raw fluorescence images were input directly to Cellpose with the ``nuclei'' pretrained model and default diameter settings. 
Segmentation masks were then compared against manually curated ground‑truth annotations on 10 fields of view. 
Overall segmentation accuracy was reported as the complement of the mean false‑positive and false‑negative rates across all test images.  

\subsection*{Assessment of Image Interpretability and Prediction Quality}
To quantitatively assess the interpretability and prediction quality of images produced by our method and baseline comparisons, a blinded questionnaire study was conducted with nine expert microscopists. 
Each participant was shown 50 fluorescent images and asked to rate three criteria: nuclear visibility (delineation of nuclear boundaries in DAPI), lumen visibility (perinuclear CD29 enrichment in MSCs), and overall image quality (artifact suppression under illumination perturbations). 
All ratings were provided on a 5-point Likert scale (1 = poor, 5 = excellent). 
Images were presented in a randomized and blinded fashion, with no information regarding imaging source, adaptation status, or prediction type disclosed. 
A brief anatomical orientation and visual reference were provided prior to the evaluation to standardize rating criteria. 
For each image, scores across participants were averaged to produce a consensus rating per criterion. 
Group-level comparisons were performed using paired t-tests to identify conditions yielding significantly different interpretability scores.

\bibliography{sample}

\section*{Acknowledgements}
This work was partially supported by NSF (CBET-2244760, DBI-2325121, IIS-2048280) and NIH NIGMS (R35GM146735). 

\section*{Author contributions statement}
K.W.K.Y, A.Bai, C.J.H, and N.Y.C.L. conceived the experiments,  A.Bermudez, Z.L., I.S., and M.L. acquired and processed the microscopy data, M.L. and V.G. constructed the early version of code., K.W.K.Y. and A.Bai finalized the coded and conducted the image adaptation. All authors analyzed the results. All authors reviewed the manuscript. 

\section*{Code and data availability}
All code required to reproduce training and adaptation, including instructions and configuration files, is openly available on \href{https://github.com/kaiwentw1018/SIT-ADDA}{GitHub}. \href{https://github.com/kaiwentw1018/SIT-ADDA}{https://github.com/kaiwentw1018/SIT-ADDA}.
Raw and processed imaging data, including MSC DAPI and CD29 images as well as cross-platform datasets, are deposited in Zenodo.
\href{https://zenodo.org/records/17066331?token=eyJhbGciOiJIUzUxMiJ9.eyJpZCI6IjIzYTBkZTJhLTdkYWMtNGQ0Yi04YjI1LTc4ZDBiYzEyZDgyOSIsImRhdGEiOnt9LCJyYW5kb20iOiIzZjE4ZjBiMGVjZTY0ZGYzNGU1YjMyZjFkMDI1NTZkNSJ9.ofMzjCSgH6EDsNEeyNyA5ARWxpszj8CTH1Pebutv4dYK7CWlPuADiuyTwCsgCi8wthzVD1p2HefWVjGLx-ErQA}{https://doi.org/10.5281/zenodo.17066331}. 

\section*{Competing Interests}
The authors declare no competing interests.

\end{document}